\documentclass[onecolumn,superscriptaddress,amsmath,amssymb,aps,showkeys,floatfix]{revtex4-2}
\usepackage{amsmath,bm}
\usepackage{amssymb}
\usepackage{amsthm}
\usepackage{amsfonts}

\usepackage{mathtools}
\usepackage{rotating}
\usepackage{capt-of}
\usepackage{array}
\usepackage{blkarray, bigstrut}
\usepackage{enumitem}
\newlist{notes}{enumerate}{1}
\setlist[notes]{label=Note: ,leftmargin=*}
\usepackage{graphicx}
\usepackage{hyperref}
\usepackage{xcolor}
\usepackage{comment}
\usepackage{makecell}

\usepackage{capt-of}
\usepackage{float}
\makeatletter
\let\newfloat\newfloat@ltx
\makeatother

\usepackage{algorithm}
\usepackage{algpseudocode}
\usepackage{float}
\usepackage{enumitem}
\usepackage[section]{placeins}
\usepackage{bm}% bold mathhttps://www.overleaf.com/project/63898e941dc3f81d3b516496
\usepackage[normalem]{ulem}
\theoremstyle{definition}

\newcolumntype{P}[1]{>{\centering\arraybackslash}p{#1}}

\usepackage{multirow}
\usepackage{booktabs}
\usepackage{rotating,tabularx}
\usepackage{ulem}
\usepackage{url}
\usepackage{hhline}
\usepackage{siunitx}
\usepackage{dcolumn}% Align table columns on decimal point

\makeatletter
\def\maketitle{
\@author@finish
\title@column\titleblock@produce
\suppressfloats[t]}
\makeatother

\makeatletter
\def\p@subsection{}
\makeatother
\makeatletter
\def\p@subsubsection{}
\makeatother

\usepackage{threeparttable}
\usepackage{listings}
\makeatletter
\renewcommand{\fnum@algorithm}{\fname@algorithm~\thealgorithm.}
\makeatother

\begin{document}
\preprint{APS}

%--------------------------------------------------------------------------
% Title
\title{Application of discrete Ricci curvature in pruning randomly wired neural networks: A case study with chest x-ray classification of COVID-19}
%--------------------------------------------------------------------------

%--------------------------------------------------------------------------
% Authors
\author{Pavithra Elumalai}
\thanks{These authors contributed equally: Pavithra Elumalai and Sudharsan Vijayaraghavan}
\affiliation{The Institute of Mathematical Sciences (IMSc), Chennai 600113, India}
\affiliation{Present Address: Institute of Computer Science and Campus Institute Data Science, University of G\"ottingen, 37073 G\"ottingen, Germany}

\author{Sudharsan Vijayaraghavan}
\thanks{These authors contributed equally: Pavithra Elumalai and Sudharsan Vijayaraghavan}
\affiliation{The Institute of Mathematical Sciences (IMSc), Chennai 600113, India}

\author{Madhumita Mondal}
\thanks{To whom correspondence should be addressed:\\ asamal@imsc.res.in or madhumitam@imsc.res.in}
\affiliation{The Institute of Mathematical Sciences (IMSc), Chennai 600113, India}
\affiliation{Homi Bhabha National Institute (HBNI), Mumbai 400094, India}

\author{Areejit Samal}
\thanks{To whom correspondence should be addressed:\\ asamal@imsc.res.in or madhumitam@imsc.res.in}
\affiliation{The Institute of Mathematical Sciences (IMSc), Chennai 600113, India}
\affiliation{Homi Bhabha National Institute (HBNI), Mumbai 400094, India}
%--------------------------------------------------------------------------

%--------------------------------------------------------------------------
% Abstract
\begin{abstract}
Randomly Wired Neural Networks (RWNNs) serve as a valuable testbed for investigating the impact of network topology in deep learning by capturing how different connectivity patterns impact both learning efficiency and model performance. 
At the same time, they provide a natural framework for exploring edge-centric network measures as tools for pruning and optimization.
In this study, we investigate three edge-centric network measures: Forman-Ricci curvature (FRC), Ollivier-Ricci curvature (ORC), and edge betweenness centrality (EBC), to compress RWNNs by selectively retaining important synapses (or edges) while pruning the rest.
As a baseline, RWNNs are trained for COVID-19 chest x-ray image classification, aiming to reduce network complexity while preserving performance in terms of accuracy, specificity, and sensitivity. 
We extend prior work on pruning RWNN using ORC by incorporating two additional edge-centric measures, FRC and EBC, across three network generators: Erd\"{o}s-R\'{e}nyi (ER) model, Watts-Strogatz (WS) model, and Barab\'{a}si-Albert (BA) model.
We provide a comparative analysis of the pruning performance of the three measures in terms of compression ratio and theoretical speedup. 
A central focus of our study is to evaluate whether FRC, which is computationally more efficient than ORC, can achieve comparable pruning effectiveness.
Along with performance evaluation, we further investigate the structural properties of the pruned networks through modularity and global efficiency, offering insights into the trade-off between modular segregation and network efficiency in compressed RWNNs.
Our results provide initial evidence that FRC-based pruning can effectively simplify RWNNs, offering significant computational advantages while maintaining performance comparable to ORC.
\end{abstract}

\keywords{randomly wired neural networks, discrete Ricci curvature, graph-based pruning, image classification, modularity, global efficiency}

%---------------------------------------------------------------
\maketitle
%---------------------------------------------------------------

%---------------------------------------------------------------
% Introduction
\section{Introduction}

Neural Architecture Search (NAS) has emerged as a powerful paradigm for automatically designing neural architectures that achieve high performance with limited computational resources and minimal human intervention \cite{Ren2021, Cheng2020, Zhang2020}. 
Inspired by the success of wiring patterns in classical deep convolutional networks (DCNs) such as ResNet \cite{He2016, Szegedy2017} and DenseNet \cite{Huang2017}, recent NAS research has explored novel and innovative wiring patterns for neural networks, aiming to eliminate human bias in network design \cite{salmani2025}. 
Among these efforts, Xie \textit{et al.} \cite{Xie2019} introduced randomly wired neural networks (RWNNs), where network connectivity is governed by three well-established random graph models in network science: the Erd\"{o}s-R\'{e}nyi (ER) model \cite{Erdos1960}, Watts-Strogatz (WS) model \cite{Watts1998}, and the Barab\'{a}si-Albert (BA) model \cite{Barabasi1999}.
These graph-based wiring schemes serve as stochastic yet structured processes for architecture generation, and RWNNs have demonstrated performance comparable to classical architectures such as ResNet and DenseNet \cite{Xie2019}. 
In this work, we extend the utility of RWNNs by exploring their application to COVID-19 classification from chest x-ray scans.
While NAS contributes to better and optimized wiring patterns, pruning offers a more traditional approach for reducing the computational demands of deep networks \cite{Blalock2020, Janowsky1989, Mozer1988, Karnin1990}. 
Pruning reduces model complexity by systematically removing parameters or connections from a large, accurate parent architecture while striving to preserve its predictive performance \cite{Blalock2020}. 
Thus, this technique can reduce computational, memory, and energy costs \cite{Molchanov2019}, and in some cases, small amounts of pruning even enhance the performance of the neural networks \cite{Han2015, Suzuki2018}.

At a more fundamental level, artificial neural networks can be represented as computational graphs in which neurons are connected by edges or links that direct the flow of data \cite{You2020}. 
Graph theory and network science thus provide a natural framework for analyzing neural networks and their behavior. 
Prior work has investigated the relationship between the accuracy of neural networks and their underlying graph structure, such as average path length and clustering coefficient \cite{You2020}, as well as how graph connectivity patterns relate to robustness against noise and adversarial attacks \cite{Waqas2022}. 
With increasing emphasis on edge-level properties in neural networks, edge-centric measures from network science offer new directions for analyzing and improving architectures. 
Edge betweenness centrality (EBC) \cite{Girvan2002} is a well-known edge-centric measure in network science and has been employed in network analysis in numerous domains. 
Graph Ricci curvatures are network geometry-based edge-centric measures that can identify the significant edges (or links) in a network based on various properties. 
Two of the most commonly used graph curvatures are Ollivier-Ricci curvature (ORC) \cite{Ollivier2007, Ollivier2009} and Forman-Ricci curvature (FRC) \cite{Forman2003, Sreejith2016, Sreejith2017} which have been successfully applied in other domains such as network finance \cite{Sandhu2016, Samal2021, kulkarni2024}, network neuroscience \cite{Farooq2019, Chatterjee2021, Elumalai2022, yadav2023}, network biology \cite{Sandhu2015, Tannenbaum2015}, and in artificial neural networks \cite{Glass2020, Waqas2022, wu2023, liu2023, han2023, shen2024}. 
Notably, Waqas \textit{et al.} \cite{Waqas2022} demonstrated that neural network robustness strongly correlates with ORC, while Shen \textit{et al.} \cite{shen2024} introduced curvature-enhanced graph convolutional networks (CGCNs) that leverage ORC to incorporate local geometric information for biomolecular interaction prediction.

A large number of studies have applied deep learning techniques for the classification of COVID-19 using chest x-ray (CXR) images. 
Classical DCNs such as ResNet \cite{He2016, Szegedy2017}, DenseNet \cite{Huang2017}, AlexNet \cite{Krizhevsky2012}, VGG \cite{Simonyan2015}, Inception \cite{Szegedy2015}, Xception \cite{Chollet2017}, and Mobilenets \cite{Howard2017} have been widely employed, often leveraging transfer learning from large-scale datasets such as ImageNet \cite{Deng2009}. 
These approaches have demonstrated strong performance in COVID-19 CXR classification tasks by repurposing well-established architectures originally designed for general image classification and other medical imaging domains \cite{yu2021, Fourcade2019}. 
However, the increasing depth and parameter complexity of such models make them computationally expensive and challenging to deploy in resource-constrained environments such as smartphones \cite{Han2015, Sze2017, Yang2017, Blalock2020, Suganyadevi2021}.

Alongside existing architectures, novel and lightweight convolutional neural network (CNN) models have also been developed specifically for COVID-19 CXR classification, such as COVID-Net \cite{WangL2020}, Covid-caps \cite{Afshar2020}, DeTraC \cite{Abbas2021}, COVIDLite \cite{Siddhartha2020}, CoroNet \cite{Khan2020}, CovXnet \cite{Mahmud2020}, Fast COVID-19 Detector (FCOD), \cite{Panahi2021}, DarkCovid net \cite{Ozturk2020}, Mobilenet with residual separable convolution block (MNRSC) \cite{Tangudu2022}, Covmnet \cite{jawahar2022}, LiteCovidNet \cite{kumar2022}, COVIDX-LwNet \cite{wang2022}, as well as custom CNN models proposed by Maghdid \textit{et al.} \cite{Maghdid2021}, Rahimzadeh \textit{et al.} \cite{Rahimzadeh2020}, Apostolopoulos \textit{et al.} \cite{Apostolopoulos2020b}, Karakanis \textit{et al.} \cite{Karakanis2021}.
These models are typically trained and evaluated on publicly available datasets such as the CovidX dataset \cite{Hemdan2020}, COVID-Xray-5k dataset \cite{Minaee2020}, Cohen dataset \cite{Cohen2020}, and ChexPert dataset \cite{Irvin2019}, addressing both binary and multi-class classification tasks. 
For a comprehensive review of these deep learning methods, we refer the reader to references \cite{Subramanian2022, siddiqui2022, ait2023}.
However, most of these works are limited to conventional CNN architectures; our focus is to explore RWNNs and pruning strategies.

By design, RWNNs offer high scope for edge-centric network measures for pruning (or compression) and optimization. 
In this paper, we used three edge-centric network measures, namely, edge betweenness centrality (EBC), Ollivier-Ricci curvature (ORC), and Forman-Ricci curvature (FRC), to compress randomly wired neural networks by identifying and retaining important synapses (or edges) and pruning the rest. 
As a baseline, we trained RWNNs for the task of COVID-19 CXR image classification. 
Our primary objective is to compress these networks while preserving their initial performance. 
Additionally, we provide a comparative analysis of the pruning performance of the three edge-centric network measures based on the compression ratio and theoretical speedup. 
While Glass \textit{et al.} \cite{Glass2020} demonstrated the use of ORC for pruning RWNNs in RicciNets with the WS model, we extend this line of work by considering all three network generators (ER, WS, and BA) and by incorporating EBC and FRC alongside ORC. 
A key focus of our study is to examine whether FRC, which is computationally more efficient than ORC, can serve as an effective alternative while maintaining comparable pruning performance. 
Finally, we investigated the pruned network structures in terms of their modularity and global efficiency.
%---------------------------------------------------------------

%---------------------------------------------------------------
% Theory
\section{Background and Preliminaries}
% \section{Theoretical Background}\label{sec_Theory}

In this section, we describe the three widely-studied random graph models in network science, and thereafter, describe how they have been extended to artificial neural networks in the literature, specifically in the form of randomly wired neural networks (RWNNs). 
Alongside, we describe the different performance and complexity evaluation metrics for artificial neural networks, and pruning evaluation metrics that are used in this paper.

%---------------------------------------------------------------
\subsection{Random graph models}

A graph \textit{G(V,E)} is a set of vertices (or nodes) \textit{V}, that are connected by a set of edges (or links) \textit{E}. 
Random graphs belong to an ensemble of graphs wherein the presence of edges is governed by a probability distribution. 
The three classical models of random graphs widely-studied in network science literature are Erd\"{o}s-R\'{e}nyi model \cite{Erdos1960}, Watts-Strogatz model \cite{Watts1998}, and the Barab\'{a}si-Albert model \cite{Barabasi1999}.

\begin{enumerate}    
    \item[(a)] \textbf{Erd\"{o}s-R\'{e}nyi (ER) model:} Proposed by Paul Erd\"{o}s and Alfr\'{e}d R\'{e}nyi, an ER graph \textit{G(n,p)} is characterised by the number of vertices \textit{`n'} where every pair of these \textit{`n'} vertices is connected distinctly with probability \textit{`p'}. 
    The presence of edges in an ER graph is independent of each other.
    \item[(b)] \textbf{Watts-Strogatz (WS) model:} Duncan J. Watts and Steven Strogatz proposed a random graph model to produce graphs with high clustering and the small-world property. 
    A WS graph \textit{G(n,k,p)} is a \textit{k}-regular graph with \textit{`n'} vertices whose edges have been rewired uniformly with a probability \textit{`p’}. 
    \item[(c)] \textbf{Barab\'{a}si-Albert (BA) model:} In an attempt to mimic real-world networks, R\'{e}ka Albert and Albert-L\'{a}szl\'{o} Barab\'{a}si proposed a model that allows for the inclusion of new nodes in the network by connecting them to the existing nodes based on their degree distribution. 
    This model of graphs \textit{G(n,m)} follows a power-law degree distribution, and thus, are scale-free in nature. 
    BA graphs evolve from an initial \textit{`$m_0$'} ($m_0 > m$) nodes and randomly connected edges. 
    An incoming node gets attached to \textit{`m'} existing nodes such that nodes with higher degree are more likely to link to the new incoming nodes.
\end{enumerate}
%---------------------------------------------------------------

%---------------------------------------------------------------
\subsection{Randomly wired neural networks}\label{subsec_RWNN} 

\begin{figure*}
\centering
\includegraphics{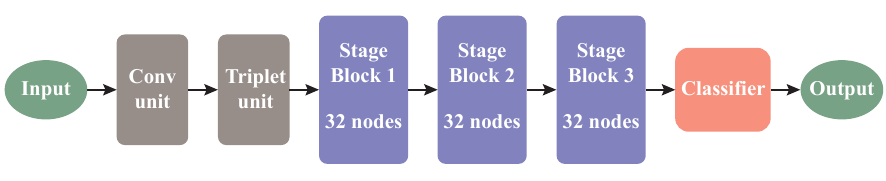}
\caption{A schematic description of the architecture of RWNNs. It comprises an input layer, one conv unit, one triplet unit, three hidden stage block layers (each with 32 nodes that are generated using one of the ER, WS or BA model with parameters as specified in table \ref{tab_networks}), a classifier layer, and an output layer, which are all are connected sequentially.}
\label{fig_HiddenLayers}
\end{figure*}

In 2019, Xie \textit{et al.} \cite{Xie2019} proposed the use of random graphs to construct artificial neural networks, specifically convolution neural networks (CNNs) for image classification, referred to as randomly wired neural networks (RWNNs). 
In their paper, Xie \textit{et al.} \cite{Xie2019} have provided a clear description of RWNN architectures and their implementation.
Similar to conventional CNNs, RWNNs consist of input, classifier, and output layers. 
Further, the input and the classifier layers sandwich a stack of hidden layer blocks.
More specifically, an RWNN comprises of an input layer, one conv unit, one triplet unit, and three hidden stage blocks with $32$ nodes (accounting for five convolutional layers), a classifier, and an output layer, which are all connected in a linear fashion (see figure \ref{fig_HiddenLayers}).

In a RWNN, each hidden layer block corresponds to a random graph that is mapped into a neural network.
The hidden layer block can be initialized by generating a random graph parameterized according to one of three classical random graph models: ER, WS, or BA.
The generated random graphs whose nodes are enumerated by numeric labels are transformed into Directed Acyclic Graphs (DAGs). 
This is achieved by adding directions to the edges in such a way that the node with the lower numerical label acts as the source and the node with the higher numerical label acts as the target. 
Two stand-alone nodes are included to act as the primary input and output nodes. 
The primary input node feeds data to the nodes in the DAG that have no incoming edges, while the primary output node collects the result from nodes in the DAG that do not have outgoing edges. 
Edges function as a medium for data flow between nodes, whereas nodes perform the following three functions: (a) \textit{aggregation}, i.e., to combine the input data from the input edges into a weighted sum; (b) \textit{transformation}, i.e., to process the aggregated data by a triplet unit composed of ReLU-convolution-BatchNorm; (c) \textit{distribution}, i.e., to distribute copies of the processed data through the output edges of a node.

In this work, we implemented ten different instances of each of the three classes of RWNNs generated using one of three classical random graph models, namely, ER, WS, or BA model and random seeds. 
The parameters used to generate each class of RWNN are listed in table \ref{tab_networks}. 
Following Xie \textit{et al.} \cite{Xie2019}, these configurations are specifically chosen given their superior performance over other choices of parameters. 
In this study, the RWNN models were implemented in Python using the PyTorch \cite{Paszke2019} library, with each instance trained for 100 epochs. 
Training was performed using stochastic gradient descent (SGD) with a learning rate of 0.1, and weight parameters were initialized from a normal distribution with zero mean and a fixed yet small standard deviation. 
For further details on RWNNs, we refer the reader to the original article by Xie \textit{et al.} \cite{Xie2019}.

%--------------------------------------------------------------------

%--------------------------------------------------------------------
% Figure 1
%--------------------------------------------------------------------

%--------------------------------------------------------------------
% Table 1
\begin{table}[!t]
\caption{The parameters used to generate random graphs using network models, namely the Erd\"{o}s-R\'{e}nyi (ER), the Watts-Strogatz (WS), and the Barab\'{a}si-Albert (BA), that were employed in the construction of RWNNs. 
These parameters were specifically chosen as per Xie \textit{et al.} \cite{Xie2019} given their superior performance in accuracy over other choices of parameters within the model class.
\label{tab_networks}}%
\begin{tabular*}{\columnwidth}{@{\extracolsep\fill}llll@{\extracolsep\fill}}
\toprule
\toprule
\textbf{Model} & \textbf{Parameters} & \textbf{Description}\\
\midrule
Erd\"{o}s-R\'{e}nyi (ER) & \textit{p $=0.2$} & $p$ denotes the probability of two nodes being connected by an edge \\
Watts-Strogatz (WS) & \textit{k $=4$, p $=0.75$} & $k$ denotes the degree of every node in a regular graph, \\ 
& & $p$ denotes the probability by which an edge is rewired from the $k$-regular graph \\
Barab\'{a}si-Albert (BA) & \textit{m $=5$} & $m$ denotes the number of new edges formed by an incoming node \\
\bottomrule
\end{tabular*}
\end{table}
%--------------------------------------------------------------------

%--------------------------------------------------------------------
\subsubsection{Performance evaluation metrics}

The performance of a trained neural network on the correctness of the output, in a classification task, can be evaluated based on many criteria. The most commonly used metrics are accuracy, specificity, sensitivity (or recall), receiver operating characteristic curve (ROC), area under this curve (AUC), precision, and F1-score. 
The accuracy is defined as the ratio of correctly classified test images (positive or negative) to the total number of images in the test set, i.e.,
\begin{equation*}
    \textit{Accuracy} = \frac{\textit{Number of images correctly classified}}{\textit{Size of test set}} 
\end{equation*}
Specificity is the ratio of the number of test images that are correctly classified as negative by the model to the total number of negative images in the test set, i.e.,
\begin{equation*}
    \textit{Specificity} = \frac{\textit{Number of images correctly classified as negative}}{\textit{Number of negative images in test set}} 
\end{equation*}
Sensitivity is the ratio of the number of test images that are correctly classified as positive by the model to the total number of positive images in the test set, i.e.,
\begin{equation*}
    \textit{Sensitivity/Recall} = \frac{\textit{Number of images correctly classified as positive}}{\textit{Number of positive images in test set}} 
\end{equation*}
The ROC curve is a plot of true positive rate (TPR) against the false positive rate (FPR) of the deep learning model at different classification thresholds, wherein TPR and FPR can be defined as follows:
\begin{equation*}
    \textit{TPR} = \frac{\textit{Number of true positives}}{\textit{(Number of true positives + Number of false negatives)}}. 
\end{equation*} 
Similarly,
\begin{equation*}
    \textit{FPR} = \frac{\textit{Number of false positives}}{\textit{(Number of false positives + Number of true negatives)}} 
\end{equation*} 
The area under this curve is referred to as AUC-ROC. 
Precision is the ratio of the number of test images that are correctly classified as positive by the model to the total number of images predicted as positive in the test set, i.e., 
\begin{equation*}
    \textit{Precision} = \frac{\textit{Number of images correctly classified as positive}}{\textit{Number of images classified as positive in test set}} 
\end{equation*}
F1-score is the harmonic mean of precision and sensitivity, providing a balance between the two measures. i.e.,
\begin{equation*}
    \textit{F1-score} = \frac{2 \times Precision \times Sensitivity}{Precision + Sensitivity}
\end{equation*}
In this study, we have used the above-mentioned six metrics namely, accuracy, specificity, sensitivity/recall, AUC-ROC, precision and F1-score to evaluate the performance of RWNNs.
%--------------------------------------------------------------------

%--------------------------------------------------------------------
\subsubsection{Complexity evaluation metrics}

The complexity of a neural network can be accounted for by its number of floating-point operations (FLOPs), also known as add-multiply operations, and the number of parameters the model comprises. 
Parameters refer to the trainable weights in the model, which are learnt by optimising a loss function. 
These parameters enable the model to generate predictions for any given input.
The number of operations for any instance of an input to achieve an output is quantitatively measured as the FLOPs. 
The higher the number of parameters and the FLOPs, the greater the computational complexity. 
It is to be noted that the number of parameters and FLOPs increase with the addition of layers to a neural network architecture.

%--------------------------------------------------------------------

%-----------------------------------------------------------------------
\subsection{Pruning of artificial neural networks}

In literature, there are several methods for pruning artificial neural networks. 
Pruning methods can be categorized based on when to prune (i.e., pre- or mid- or post-training) and which parameters to prune for ease of understanding. 
Considering when to prune a neural network, some methods propose pruning during network initialization \cite{Wang2021b, Lee2018}, while other methods propose pruning periodically amidst training \cite{Gale2019}. 
Notably, most of the methods prune the network after training \cite{Han2015}.
Parameters targeted for pruning can be removed in a single step, by eliminating all at once \cite{Liu2018}, or iteratively, by removing a fixed fraction at each step \cite{Han2015}, or adaptively, by removing a varying fraction across successive steps \cite{Gale2019}. 
Based on which parameters to remove from the neural networks, pruning methods can be unstructured or structured. 
Unstructured pruning makes the network sparse by pruning individual parameters in the network. 
In contrast, structured pruning considers groups of parameters such as entire neurons, filters, or channels, which are structural elements in the network for removal. 
The parameters to remove could be selected based on their absolute values, trained importance coefficients, or their significance towards network activation and gradients \cite{Blalock2020}. 
Methods that involve pruning at initialization, train the neural network right after pruning. 
Methods that involve pruning post-training could further train the model to recover from any loss in performance \cite{Han2015}, or rewind to an earlier state \cite{Frankle2018}, or reinitialize \cite{Liu2018} to train from scratch again. 
We refer the reader to Blalock \textit{et al.} \cite{Blalock2020} for a detailed review of the pruning methods in literature.
%-----------------------------------------------------------------------

%-----------------------------------------------------------------------
\subsubsection{Pruning evaluation metrics}

In literature, the commonly used metrics to evaluate pruning are compression ratio and theoretical speedup. 
The compression ratio is defined as the ratio of the size of the original network to the size of the new pruned state of the network, i.e.,
\begin{equation*}
    \textit{Compression ratio} = \frac{\textit{Number of parameters in original network}}{\textit{Number of parameters in pruned network}}.
\end{equation*} 
Similarly, theoretical speedup is defined as the number of FLOPs in the original network to the number of FLOPs in the pruned state of the network, i.e.,
\begin{equation*}
    \textit{Theoretical speedup} = \frac{\textit{Number of FLOPs in original network}}{\textit{Number of FLOPs in pruned network}}.
\end{equation*} 

Blalock \textit{et al.} \cite{Blalock2020} recommend the use of both metrics to report the results of pruning, and we follow their suggestion in this work. 
For ease of inference, we also report the results in terms of the percentage of the parameters and FLOPs reduced after pruning.
%---------------------------------------------------------------

%---------------------------------------------------------------
\subsection{Edge-centric network measures}\label{subsec_EdgeBasedNM}

Most of the widely-used measures in network science, such as degree, clustering coefficient, and betweenness centrality, are node-centric. 
In other words, such measures are defined for a node in the network. 
In contrast, there are fewer measures which are edge-centric. 
Edge-centric network measures enable us to evaluate the interaction between a pair of nodes (i.e., an edge) in a network. 
In this work, we consider three edge-centric network measures to identify and prune insignificant edges from the RWNNs.

First, we employ the measure, edge betweenness centrality (EBC) \cite{Freeman1977, Girvan2002}, which can be used to quantify the importance of an edge for the flow of information globally in the network.
For every edge $e\in E$ in a graph $G(V,E)$, EBC is defined as:
\begin{equation*}
C_{EB}(e) = \sum\limits_{i,j\in V} \frac{\sigma(i,j | e)}{\sigma(i,j)}
\end{equation*}
where $V$ is the set of nodes, $\sigma(i,j)$ is the number of all the shortest paths, and $\sigma(i,j|e)$ is the number of shortest paths that pass through the edge $e$. 

In addition to the EBC, we consider two discretizations of Ricci curvature, namely Ollivier-Ricci curvature (ORC) and Forman-Ricci curvature (FRC), as edge-centric measures in this work, which are described in the next section.
%---------------------------------------------------------------

%---------------------------------------------------------------
\subsection{Graph Ricci curvatures}\label{subsec_Ricci}

Curvature is the measure of deviation of a space from being flat. 
The notion of Ricci curvatures was originally defined for smooth manifolds, capturing two of their important geometric properties, specifically, volume growth and dispersion of geodesics. 
In order to apply to networks and graphs, classical Ricci curvature has to be discretized. When discretizing, curvature is naturally assigned to edges since the classical notion is associated with a vector (direction) \cite{Jost2017}. 
Even when discretizations of the classical Ricci curvature retain some key properties, they do not retain all of the properties \cite{Samal2018}. 
It is important to note that different discretizations capture different properties. 
Simply stated, discrete notions of Ricci curvature assign a value to an edge, but the value assigned is based on different properties of the network by different discretizations. 
In this study, we have used two established notions of discrete Ricci curvatures, namely, Ollivier-Ricci curvature (ORC) and Forman-Ricci curvature (FRC). 
ORC captures the volumetric growth of networks while FRC provides insights on information spreading across the network \cite{Samal2018}. 
%---------------------------------------------------------------

%---------------------------------------------------------------
\subsubsection{Ollivier-Ricci curvature (ORC)}
 
In spaces with positive curvature, two balls (volumes) tend to be, on average, closer to each other than the distance between their centers. 
Conversely, in spaces with negative curvature, they are, on average, farther apart than the distance between their centers.
Based on this observation, Ollivier defined his discretization of the classical Ricci curvature \cite{Ollivier2007, Ollivier2009}, extending it from balls (volumes) to measure probabilities. 
Instead of a ball of radius $\epsilon$ centered at $x$, Ollivier's discretization of the classical Ricci curvature used an arbitrary probability measure around $x$. 
The ORC of an edge $e$ between nodes $i$ and $j$ in graph $G$ is defined as: 
\begin{equation*}
  \mathbf{O}(e)  = 1 - \frac{W_1(m_i,m_j)}{d(i,j)}\ ,
\end{equation*}
\noindent where $m_i$ and $m_j$ are the discrete probability measures defined on nodes $i$ and $j$, respectively, and $d(i,j)$ is the distance between $i$ and $j$. 
For an unweighted graph, $d(i,j)$ is defined as the number of edges contained in the shortest path connecting $i$ and $j$.
$W_1$ denotes the Wasserstein distance \cite{Vaserstein1969}, which is the trasportation distance between $m_i$ and $m_j$, given by:
\begin{equation*}
W_1(m_i, m_j)=\inf_{\mu_{i,j}\in \prod(m_i, m_j)}\sum_{(i',j')\in V\times V}
d(i', j')\mu_{i,j}(i', j'),
\end{equation*}
\noindent where $\prod(m_i, m_j)$ is the set of probability measures $\mu_{i,j}$ that satisfy:
\begin{equation*}
\sum_{j'\in V}\mu_{i,j}(i', j')=m_i(i'), \,\,\sum_{i'\in V}\mu_{i,j}(i', j') = m_j(j').
\end{equation*}
\noindent The above equation gives all the possible transportations of measure $m_i$ to $m_j$, and the Wasserstein distance $W_1(m_i, m_j)$ is the minimal cost of transporting $m_i$ to $m_j$. 
Note that the probability distribution $m_i$ is specified beforehand, and in our work, it is chosen to be uniform over the neighboring nodes of $i$ \cite{Lin2011}.

%---------------------------------------------------------------

%---------------------------------------------------------------
\subsubsection{Forman-Ricci curvature (FRC)}

Forman’s discretization of the classical Ricci curvature \cite{Forman2003} was later extended and defined in the context of complex networks \cite{Samal2018, Sreejith2016, Sreejith2017}. 
FRC of an edge measures the information spread at the ends of edges in a network. 
A more negative value of FRC for an edge indicates that the information spread across that edge is higher. 
For an edge $e$ between nodes $i$ and $j$ in the graph $G$, FRC is defined as: 
\begin{equation*}
\nonumber
\mathbf{F}(e) = w_e \left( \frac{w_{i}}{w_e} +  \frac{w_{j}}{w_e}  
- \sum_{e_{i}\ \sim\ e,\ e_{j}\ \sim\ e} 
\left[\frac{w_{i}}{\sqrt{w_e w_{e_{j} }}} 
+ \frac{w_{j}}{\sqrt{w_e w_{e_{j} }}} \right] \right)\,
\end{equation*}
\noindent where $w_e$ denotes the weight of the edge $e$, $w_{i}$ and $w_{j}$ denote the weights associated with the nodes $i$ and $j$, respectively, $e_{i} \sim e$ and $e_{j} \sim e$ denote the set of edges incident on nodes $i$ and $j$, respectively, after excluding the edge $e$. 
For an unweighted graph, all nodes and edges in $G$ are assigned a weight equal to $1$. 
Thus, the expression for FRC reduces to:
\begin{equation*}
\mathbf{F}(e) =   4 - deg(i) - deg(j)
\end{equation*}
\noindent where $deg(i)$ and $deg(j)$ are the degrees of nodes $i$ and $j$, respectively.

%---------------------------------------------------------------

%---------------------------------------------------------------
% Table 2
\begin{table}
\caption{The train/test split of the COVID-Xray-5k image dataset \cite{Minaee2020}. 
The x-ray images of patients with COVID-19 form the positive data class. 
The images of patients with no findings or respiratory illnesses other than COVID-19 form the negative data class. 
The number of images in the positive class training sample was increased by five times using image augmentation.}
\label{tab_Covdata}
\begin{tabular*}{0.6\columnwidth}{@{\extracolsep\fill}lccc@{\extracolsep\fill}}
\toprule
\toprule
\textbf{Split} & \textbf{Non-COVID (negative)} & \textbf{COVID (positive)}\\
\midrule
Train & 2000 & 84 original; 420 augmented\\
Test & 3000 & 100\\
\bottomrule
\end{tabular*}
\end{table}
%---------------------------------------------------------------

%---------------------------------------------------------------
% Table 3
\begin{table}
\caption{The transformation of the input images as they pass through every layer of the RWNN with $N$ number of nodes and $C$ channel count. 
The dimensions of the images change due to the stride in convolutions in every stage. 
All the input images have an initial dimension of $224\times 224$.} 
\label{tab_config}
\begin{minipage}{0.7\columnwidth}
\begin{tabular*}{\columnwidth}{@{\extracolsep\fill}lccc@{\extracolsep\fill}}
\toprule
\toprule
\textbf{Stage} & \textbf{Stage configurations$^{(a)}$} & \textbf{Output}\\
\midrule
conv1 & \textit{$3*3conv$, C} & $112*112$\\
conv2 & \textit{$3*3conv$, C} & $56*56$\\
conv3 & \textit{N, C} & $28*28$\\
conv4 & \textit{N, 2C} & $14*14$\\
conv5 & \textit{N, 4C} & $7*7$\\
classifier & classifier configs.$^{(b)}$ & $1*1$\\
\bottomrule
\end{tabular*}
\begin{tablenotes}
    \item $^{(a)}$ $N = 32$ and $C = 78$ as defined by Xie \textit{et al.} \cite{Xie2019}.
    \item $^{(b)}$ classifier configs.: $1\times 1$ conv, $1280$-d global average pool, $1000$-d fc, softmax
\end{tablenotes}
\end{minipage}
\end{table}
%---------------------------------------------------------------

%---------------------------------------------------------------
% Dataset
\subsection{COVID-19 chest x-ray dataset}%{Dataset}
\label{sec_dataset}

In this study, we used the \textit{COVID-Xray-5k} dataset curated and made available by Minaee \textit{et al.} \cite{Minaee2020}. 
The dataset contains $184$ anterior-posterior chest x-ray images of patients affected by COVID$-19$, collected from the \textit{covid-Chestxray-dataset}. 
These images form the positive training/test sample and have been verified by a board-certified radiologist for a positive diagnosis of COVID$-19$ according to Minaee \textit{et al.} \cite{Minaee2020}. 
In addition, the dataset consists of $5000$ anterior-posterior chest x-ray images of patients with no findings or respiratory conditions other than COVID$-19$, such as Pneumonia or Edema, that affect the lungs. 
These $5000$ images have been collected from the \textit{ChexPert} dataset \cite{Irvin2019} and \textit{covid-Chestxray-dataset} \cite{Cohen2020}, and they form the negative training/test sample.

The train/test split of the dataset is summarized in table \ref{tab_Covdata}. 
Since the number of positive image samples in the training set is relatively small, we increased the number of images by augmenting the original sample of images. 
The augmentation was performed in a similar manner to Minaee \textit{et al.} \cite{Minaee2020}.
Image augmentation is the process of generating new and artificial images that resemble an original pool of images while preserving data labels \cite{Bloice2017}. 
Image augmentation was performed using the Python library \textit{Augmentor} \cite{Bloice2017} to increase the number of images in the positive training sample by five times (see table \ref{tab_Covdata}). 
Importantly, all the images in both train and test splits were transformed to a uniform size of $224 \times 224$ pixels.
Lastly, for this \textit{COVID-Xray-5k} dataset considered in this study, the transformation of the input images ($224 \times 224$) as they pass through every layer in the RWNN is shown in table \ref{tab_config}.

%---------------------------------------------------------------

%---------------------------------------------------------------
% Results and Discussion
\section{Results and Discussion}

In this study, we constructed three classes of RWNNs with three classical random graph models of Erd\"{o}s-R\'{e}nyi (ER) model, Watts-Strogatz (WS) model, and the Barab\'{a}si-Albert (BA) model as specified by Xie \textit{et al.} \cite{Xie2019} (see section \ref{subsec_RWNN}). 
Corresponding to each class of RWNNs, we constructed ten different instances of RWNNs using different random seeds. 
Notably, the computations were performed mainly using a Google Colab Pro, and due to limited computational resources, we restricted the number of repeated trials to ten per class.
Subsequently, we trained the three classes of RWNNs on the \textit{COVID-Xray-5k dataset} \cite{Minaee2020} and evaluated their test performance using metrics such as accuracy, specificity, sensitivity/recall, AUC-ROC, precision, and F1-score. 
Next, considering this performance as the baseline, we pruned the networks by removing edges from the underlying graph structure according to edge-centric network measures, namely FRC, ORC, and EBC (see section \ref{subsec_pruneedge}). 
The primary goal of our study is to investigate the pruning potential of three edge-centric network measures across RWNNs corresponding to three classes of random graphs. 
To achieve this objective, we performed a binary search-based approach and removed $x$ fraction of edges (synapses) from the network at consecutive steps of the binary search. 
The search was performed up to a depth of 5 to arrive at the ensuing results.
This process yields a compressed version of the original network that maintains, or potentially improves, the performance.

%---------------------------------------------------------------

%---------------------------------------------------------------
\subsection{Performance of RWNNs in classification of CXR images of COVID-19}

We evaluated the performance of the RWNNs in terms of six metrics, namely accuracy, specificity, recall or sensitivity, AUC-ROC, precision, and F1-score. 
Our models were trained on an imbalanced but augmented dataset, which necessitated evaluating performance with sensitivity and F1-score in addition to accuracy, to appropriately capture performance.
Since we considered ten different instances for each class of RWNNs, we reported the results of these metrics as percentages, presenting both the average and maximum values together with box plots (see figure \ref{fig_Measures_comparison} and table \ref{tab_unpruned}).
Additionally, table \ref{tab_confusion} provides an overall summary of the confusion matrix components across all ten different instances.

Firstly, we found that the mean accuracy of the ER class of RWNNs over ten different instances is 96.848. 
Similarly, the mean accuracy of the WS class of RWNNs is 96.852, while that of the BA class is 96.79. 
The ER class achieved both the highest accuracy of 97.387 and the lowest accuracy of 96. 
Among the three classes, the WS class demonstrated the highest average accuracy, marginally outperforming both the BA and ER classes.

Secondly, we observed that the mean specificity for the ER class of RWNNs is 97.137, while the WS and BA classes have mean specificities of 97.147 and 97.09, respectively. 
Notably, the ER class achieved both the maximum specificity of 97.633 and the minimum specificity of 96.23.
On comparison across all three classes, the WS class has a higher mean specificity for the classification of the CXR images with COVID-19. 

Thirdly, the mean sensitivities (recall) of the ER, WS, and BA classes were 88.2, 88.0, and 87.8, respectively. 
The BA class achieved the maximum sensitivity of 92, while the ER class recorded the lowest at 83. 
Despite this variation, the ER class achieved the highest average sensitivity across trials. 
This relatively high recall reflects the model’s ability to identify COVID-19 positive cases effectively, which is critical in a screening setting where minimizing false negatives is more important than avoiding some false positives.

Fourthly, we evaluated performance using the AUC-ROC metric for CXR image classification for COVID-19. 
On average, the ER class achieved the highest AUC-ROC of 97.89, compared to 96.399 for WS and 96.643 for BA. 
While the BA class achieved the overall maximum AUC-ROC of 98.561 and the WS class the minimum of 94.927, the ER class outperformed the others in terms of average AUC-ROC.

Fifthly, precision values were lower due to dataset imbalance, with the ER, WS, and BA classes averaging 50.844, 50.8, and 50.207, respectively.
The ER class achieved the highest precision of 55.901 but also the lowest of 44.059. 
Importantly, although precision was modest, this is an expected outcome in a low-prevalence setting with far more negative than positive cases. 
In clinical screening, prioritizing recall ensures fewer missed COVID-19 patients, while flagged false positives can be resolved with confirmatory testing.

Finally, the F1-scores reflect the balance between precision and recall.
The ER, WS, and BA classes attained mean F1-scores of 64.444, 64.384, and 63.861, respectively. 
The ER class not only achieved the highest F1-score (68.966) but also showed the widest range, from 58.94 to 68.966. 
Overall, the ER class demonstrated the most balanced performance across metrics, with consistently high recall and competitive F1-scores, making it particularly suitable for COVID-19 screening tasks where recall is paramount.

Figures \ref{fig_Measures_comparison}(a)-(f) display the box plots (in gray) of accuracy, specificity, sensitivity, AUC-ROC, precision, and F1-score obtained for the three RWNN classes across ten instances. 
Table \ref{tab_unpruned} summarizes the average and maximum performance of each class, reported as percentages, across six evaluation matrices.

During initialization, to measure the complexities of the RWNNs, we considered two metrics: (i) the number of parameters, and (ii) the number of floating-point operations (FLOPs), which is also known as the number of add-multiply operations. 
The BA class exhibited the highest average FLOPs and parameter count, whereas the WS class demonstrated the lowest averages for both measures.
The complexity values for the RWNNs are summarized in table \ref{tab_flopsParam}. 

%---------------------------------------------------------------

%---------------------------------------------------------------
% Table 4
\begin{table}
\caption{Performance summary for the three classes of RWNNs in terms of accuracy, specificity, sensitivity, AUC-ROC, precision, and F1-score.}
\label{tab_unpruned}
\begin{tabular*}{\columnwidth}
{@{\extracolsep\fill}lccccccccccccc@{\extracolsep\fill}}
\toprule 
\toprule
\multirow{2}{*}{\textbf{Model}} & \multicolumn{2}{c}{\textbf{Accuracy (\%)}} & \multicolumn{2}{c}{\textbf{Specificity (\%)}} & \multicolumn{2}{c}{\textbf{Sensitivity (\%)}} & \multicolumn{2}{c}{\textbf{AUC-ROC (\%)}} & \multicolumn{2}{c}{\textbf{Precision (\%)}} & \multicolumn{2}{c}{\textbf{F1-score (\%)}} \\
\cline{2-3} \cline{4-5} \cline{6-7} \cline{8-9} \cline{10-11} \cline{12-13}
& Average & Max & Average & Max & Average & Max & Average & Max & Average & Max & Average & Max\\
\midrule
\textbf{ER} & 96.848 & 97.387 & 97.137 & 97.633 & 88.2 & 91 & 97.89 & 98.51 & 50.844 & 55.901 & 64.444 & 68.966 \\
\textbf{WS} & 96.852 & 97.258 & 97.147 & 97.567 & 88 & 91 & 96.399 & 97.177 & 50.8 & 54.658 & 64.384 & 67.433 \\
\textbf{BA} & 96.79 & 97.065 & 97.09 & 97.367 & 87.8 & 92 & 96.643 & 98.561 & 50.207 & 52.695 & 63.861 & 66.667 \\
\bottomrule
\end{tabular*}
\end{table}
%---------------------------------------------------------------

%---------------------------------------------------------------
% Table 5
\begin{table}
\caption{Summary of confusion matrix components for the three classes of RWNNs, namely, Erd\"{o}s-R\'{e}nyi (ER) model, Watts-Strogatz (WS) model, and  Barab\'{a}si-Albert (BA) model, under four scenarios: before pruning (unpruned) and after pruning based on three edge-centric network measures, FRC, ORC, and EBC. The table reports True Positives (TP), True Negatives (TN), False Positives (FP), and False Negatives (FN) for each of the configuration across 10 random seeds.}
\label{tab_confusion}
\begin{tabular*}{\columnwidth}
{@{\extracolsep\fill}ccccccccccccccc@{\extracolsep\fill}}
\toprule 
\toprule 
\multirow{2}{*}{\shortstack[c]{\textbf{Random}\\ \textbf{seed}}} & \multirow{2}{*}{\shortstack[c]{\textbf{Edge}\\ \textbf{measure}}} & \multicolumn{4}{c}{\textbf{ER}} & \multicolumn{4}{c}{\textbf{WS}} & \multicolumn{4}{c}{\textbf{BA}} \\
\cline{3-6} \cline{7-10} \cline{11-14}
& & TP & TN & FP & FN & TP & TN & FP & FN & TP & TN & FP & FN \\
\midrule
\multirow{4}{*}{\textbf{3}} & Unpruned & 90 & 2929 & 71 & 10 & 87 & 2898 & 102 & 13 & 86 & 2914 & 86 & 14 \\
 & FRC & 91 & 2930 & 70 & 9 & 87 & 2903 & 97 & 13 & 87 & 2919 & 81 & 13 \\
 & ORC & 90 & 2929 & 71 & 10 & 92 & 2904 & 96 & 8 & 86 & 2944 & 56 & 14 \\
 & EBC & 91 & 2930 & 70 & 9 & 89 & 2906 & 94 & 11 & 88 & 2936 & 64 & 12 \\
 \midrule
\multirow{4}{*}{\textbf{16}} & Unpruned & 88 & 2901 & 99 & 12 & 89 & 2922 & 78 & 11 & 88 & 2917 & 83 & 12 \\
 & FRC & 88 & 2901 & 99 & 12 & 89 & 2922 & 78 & 11 & 90 & 2925 & 75 & 10 \\
 & ORC & 88 & 2901 & 99 & 12 & 90 & 2925 & 75 & 10 & 88 & 2933 & 67 & 12 \\
 & EBC & 88 & 2901 & 99 & 12 & 90 & 2929 & 71 & 10 & 90 & 2919 & 81 & 10 \\
 \midrule
\multirow{4}{*}{\textbf{34}} & Unpruned & 89 & 2915 & 85 & 11 & 86 & 2911 & 89 & 14 & 91 & 2918 & 82 & 9 \\
 & FRC & 89 & 2920 & 80 & 11 & 87 & 2911 & 89 & 13 & 91 & 2918 & 82 & 9 \\
 & ORC & 89 & 2915 & 85 & 11 & 86 & 2911 & 89 & 14 & 91 & 2918 & 82 & 9 \\
 & EBC & 89 & 2915 & 85 & 11 & 87 & 2920 & 80 & 13 & 91 & 2918 & 82 & 9 \\
 \midrule
\multirow{4}{*}{\textbf{57}} & Unpruned & 83 & 2916 & 84 & 17 & 86 & 2920 & 80 & 14 & 85 & 2914 & 86 & 15 \\
 & FRC & 86 & 2942 & 58 & 14 & 88 & 2938 & 62 & 12 & 88 & 2916 & 84 & 12 \\
 & ORC & 89 & 2943 & 57 & 11 & 87 & 2931 & 69 & 13 & 89 & 2929 & 71 & 11 \\
 & EBC & 87 & 2929 & 71 & 13 & 86 & 2923 & 77 & 14 & 89 & 2942 & 58 & 11 \\
 \midrule
\multirow{4}{*}{\textbf{59}} & Unpruned & 89 & 2918 & 82 & 11 & 91 & 2917 & 83 & 9 & 92 & 2908 & 92 & 8 \\
 & FRC & 92 & 2926 & 74 & 8 & 91 & 2917 & 83 & 9 & 92 & 2908 & 92 & 8 \\
 & ORC & 89 & 2918 & 82 & 11 & 91 & 2929 & 71 & 9 & 92 & 2908 & 92 & 8 \\
 & EBC & 90 & 2919 & 81 & 10 & 91 & 2917 & 83 & 9 & 92 & 2908 & 92 & 8 \\
 \midrule
\multirow{4}{*}{\textbf{61}} & Unpruned & 89 & 2887 & 113 & 11 & 88 & 2927 & 73 & 12 & 88 & 2921 & 79 & 12 \\
 & FRC & 89 & 2916 & 84 & 11 & 88 & 2927 & 73 & 12 & 88 & 2927 & 73 & 12 \\
 & ORC & 89 & 2887 & 113 & 11 & 89 & 2943 & 57 & 11 & 88 & 2921 & 79 & 12 \\
 & EBC & 90 & 2910 & 90 & 10 & 89 & 2932 & 68 & 11 & 88 & 2921 & 79 & 12 \\
 \midrule
\multirow{4}{*}{\textbf{66}} & Unpruned & 90 & 2918 & 82 & 10 & 88 & 2921 & 79 & 12 & 88 & 2897 & 103 & 12 \\
 & FRC & 90 & 2918 & 82 & 10 & 88 & 2925 & 75 & 12 & 89 & 2918 & 82 & 11 \\
 & ORC & 90 & 2918 & 82 & 10 & 88 & 2943 & 57 & 12 & 92 & 2915 & 85 & 8 \\
 & EBC & 90 & 2918 & 82 & 10 & 90 & 2923 & 77 & 10 & 92 & 2897 & 103 & 8 \\
 \midrule
\multirow{4}{*}{\textbf{72}} & Unpruned & 83 & 2922 & 78 & 17 & 87 & 2907 & 93 & 13 & 87 & 2917 & 83 & 13 \\
 & FRC & 84 & 2933 & 67 & 16 & 91 & 2929 & 71 & 9 & 89 & 2919 & 81 & 11 \\
 & ORC & 87 & 2923 & 77 & 13 & 88 & 2944 & 56 & 12 & 87 & 2917 & 83 & 13 \\
 & EBC & 87 & 2930 & 70 & 13 & 88 & 2923 & 77 & 12 & 90 & 2930 & 70 & 10 \\
 \midrule
\multirow{4}{*}{\textbf{92}} & Unpruned & 90 & 2920 & 80 & 10 & 88 & 2908 & 92 & 12 & 88 & 2915 & 85 & 12 \\
 & FRC & 90 & 2920 & 80 & 10 & 91 & 2919 & 81 & 9 & 88 & 2915 & 85 & 12 \\
 & ORC & 90 & 2928 & 72 & 10 & 92 & 2935 & 65 & 8 & 91 & 2934 & 66 & 9 \\
 & EBC & 92 & 2924 & 76 & 8 & 89 & 2930 & 70 & 11 & 90 & 2915 & 85 & 10 \\
 \midrule
\multirow{4}{*}{\textbf{97}} & Unpruned & 91 & 2915 & 85 & 9 & 90 & 2913 & 87 & 10 & 85 & 2906 & 94 & 15 \\
 & FRC & 91 & 2915 & 85 & 9 & 90 & 2921 & 79 & 10 & 86 & 2925 & 75 & 14 \\
 & ORC & 91 & 2915 & 85 & 9 & 90 & 2913 & 87 & 10 & 90 & 2929 & 71 & 10 \\
 & EBC & 91 & 2915 & 85 & 9 & 90 & 2913 & 87 & 10 & 87 & 2920 & 80 & 13 \\
\bottomrule
\end{tabular*}
\end{table}
%---------------------------------------------------------------

%---------------------------------------------------------------
% Table 6
\begin{table}
\caption{The average number of floating-point operations (FLOPs) and the average number of parameters across the ten different instances constructed for each of the three different classes of RWNNs, namely ER model, WS model, and BA model.}
\label{tab_flopsParam}
\begin{tabular*}{0.6\columnwidth}{@{\extracolsep\fill}lccc@{\extracolsep\fill}}
\toprule
\toprule
\textbf{Model} & \textbf{Baseline Parameters} & \textbf{Baseline FLOPs}\\
\midrule
\textbf{ER} & 4,364,300 & 2,041,517,235\\
\textbf{WS} & 4,346,196 & 2,038,300,640\\
\textbf{BA} & 4,662,096 & 2,133,697,760\\
\bottomrule
\end{tabular*}
\end{table}
%---------------------------------------------------------------

%---------------------------------------------------------------
% RWNN pruning
\subsection{RWNN pruning based on edge-centric measures}\label{subsec_pruneedge}
 
In this subsection, we describe our method based on edge-centric measures to prune RWNNs, aiming to obtain the smallest possible versions of individual RWNNs by identifying and retaining the most salient computational paths and pruning out insignificant edges. 
Importantly, we attempt to achieve smaller configurations of individual RWNNs without loss in performance over their original unpruned version. 
We also compare the pruning performance of edge-centric measures across the three different classes of RWNNs, namely, ER, WS, and BA.

To begin with, we consider the initial unpruned RWNN structure to be the smallest version. 
With a binary search approach, we aim to determine the optimal fraction of edges (or synapses) ranging from $0\%$ (a structure with all edges) and $100\%$ (a structure with no edges) to prune from the original network, such that the performance of the original network is not compromised.
Binary search works by considering two values, $min$ and $max$. 
The output of every iteration is the value midway between $min$ and $max$, while their values are assigned in the previous iteration based on whether the desired value we are searching for, would lie to the left or to the right of the output value on the number line. 
The initial value of $min$ is $0\%$, and the initial value of $max$ is $100\%$. 
We employ a binary search approach to determine the fraction of edges to be removed, and edge-centric network measures to identify the specific edges targeted for removal. 
Informally, edge-centric network measures such as EBC, ORC, and FRC rank the edges in a network from the most significant to the least significant by assigning a value to each of the edges in terms of the different properties captured by the measures (see sections \ref{subsec_EdgeBasedNM} and \ref{subsec_Ricci}). 
We remove a fraction of the least significant edges in every step of the binary search. 
It is to be noted that the network measure that is used (EBC, ORC, or FRC) is a hyperparameter in our pruning algorithm, and they are in no way used simultaneously. 
In fact, the three measures are compared against each other in terms of their potential to identify the edges to be pruned from RWNNs without compromising performance.

As summarized in Algorithm \ref{algo_prune}, the first step of the binary search is to remove $50\%$ (i.e., $(0\%+100\%)/2$) of the edges. 
We start by removing $50\%$ of the least significant edges from the initial model structure according to the network measure in use. 
This intermediary pruned model is trained for 100 epochs, and its performance is evaluated based on performance measures, namely, sensitivity, specificity, and accuracy. 
If the intermediary pruned model meets the baseline performance of its unpruned version, we further prune the network by removing more of the less significant edges, which in binary search terms, corresponds to $75\%$ (i.e., $(50\%+100\%)/2$). 
If the intermediary pruned model does not achieve the baseline performance, we reduce the amount of pruning by including some of the pruned edges back to the intermediary model, which in binary search terms, corresponds to $25\%$ (i.e., $(0\%+50\%)/2$). 
This process is carried out 5 times, which is $log(n)$ times, where $n$ is $32$ (number of nodes).

%---------------------------------------------------------------
% Algorithm 1
\begin{algorithm}
\caption{Algorithm to prune a graph G based on edge-centric measures}\label{algo_prune}
\begin{algorithmic}[1]
\Function{Prune}{G,edgemeasure, $acc_G$, $spec_G$, $sens_G$} \Comment{edgemeasure: edge-centric network measures (EBC/ORC/FRC)}

    \Comment{Initialize count =1, minimum = 0, and maximum = 100}
    \State count $\Leftarrow$ 1
    \State $min \Leftarrow 0$
    \State $max \Leftarrow 100$
    \State BestConfig $\Leftarrow config_G$
    \Comment{Initialize configuration of G as best configuration}
    % \State EdgeMeasures = CalculateEdgeMeasures(G, edgemeasure)
    \While{count $\leq 5$}
        \State prunePerc $\Leftarrow$ (min+max)/2
        \State SortedEdges $\Leftarrow$ Sort edges by edgemeasure
        \State LeastSignificantEdges $\Leftarrow$ find least significant edges(SortedEdges, pruned percent)
        
        \Comment{Prune prunePerc\% of edges from G, those are least significant}
        \State P $\Leftarrow$ G - LeastSignificantEdges % {least significant prunePerc no. of edges}$
        \State $config_P$ $\Leftarrow$ Train(P)
        \State $acc_P, spec_P, sens_P \Leftarrow$ Eval($config_P$)
        \If{$acc_P\geq acc_G \And spec_P\geq spec_G \And sens_P\geq sens_G$}
            \State $min \Leftarrow$ prunePerc
            \State BestConfig $\Leftarrow config_P$
        \Else
            \State $max \Leftarrow$ prunePerc
        \EndIf
        \State count $\Leftarrow$ count + 1
    \EndWhile
    \State \Return{BestConfig}
\EndFunction
\end{algorithmic}
\end{algorithm}
%---------------------------------------------------------------

It is to be noted that, when pruning edges, a few nodes could become isolated. 
We remove the isolated nodes from the network. 
If the removal of edges disconnects the graph, we connect the individual components to the primary input node of a hidden layer block, which ensures data flow and computation. 
The process of constructing and pruning an RWNN is summarized in the pseudocode Algorithm \ref{algo_pseudo}.

%---------------------------------------------------------------
\begin{algorithm}
\caption{Pseudocode summarizing the process of constructing and pruning a RWNN}\label{algo_pseudo}
\begin{algorithmic}[1]
\State G $=$ GenerateRandomGraph() \Comment{Generate random graph from three classical random graph models: ER/WS/BA}
\State $config_G$ $=$ constructRWNN(G)
\Comment{Construct RWNN from the random graphs: ER/WS/BA}
\State $acc_G$, $spec_G$, $sens_G$ $=$ Eval($config_G$)
\Comment{Evaluate the accuracy, specificity, and sensitivity of that configuration}
\State edgemeasure $=$ input() 
\Comment{Calculate the edge-centric measures: EBC/ORC/FRC}
\State CalcEdgeMeas(G, edgemeasure)
\State prunedModel $=$ PRUNE(G, edgemeasure, $acc_G$, $spec_G$, $sens_G$)
\end{algorithmic}
\end{algorithm}
%---------------------------------------------------------------

Thus, we prune the network at initialization before training. 
At each step, a decision is made whether the network should be pruned further or not. 
If not, we would reduce the amount of pruning in the next iteration. 
During every iteration, we rewind to the initial state of the network in terms of parameter values before training the new (intermediary or final) pruned version of RWNN. 
We emphasize that we do not consider the values or significance of individual parameters or groups of parameters. 
Instead, we prune structural elements of the network (synapses or isolated neurons) solely based on the properties of the underlying graph structure as captured by the edge-centric network measures. 

%---------------------------------------------------------------

%---------------------------------------------------------------
\subsection{Performance of RWNNs after pruning based on edge-centric network measures}

We investigated the pruning potential of three edge-centric network measures: EBC, ORC, and FRC. 
To evaluate their effectiveness, we employed a binary search strategy (Algorithm \ref{algo_prune}), in which a fraction $x$ of edges was removed from the random graph structure of the RWNNs. 
Each pruned model was then trained for 100 epochs using stochastic gradient descent (SGD) with a learning rate of 0.1, and weight parameters were initialized from a normal distribution with mean 0 and a small constant standard deviation.
The performance of the pruned models was compared with that of the original (unpruned) models using accuracy, specificity, and sensitivity, based on which the binary search determines whether to prune additional edges or restore some.
The process was carried out up to a depth of five, enabling us to identify the smallest configuration of the original network that maintains equivalent performance. 

\begin{figure*}
\centering
\includegraphics[width=0.95\linewidth]{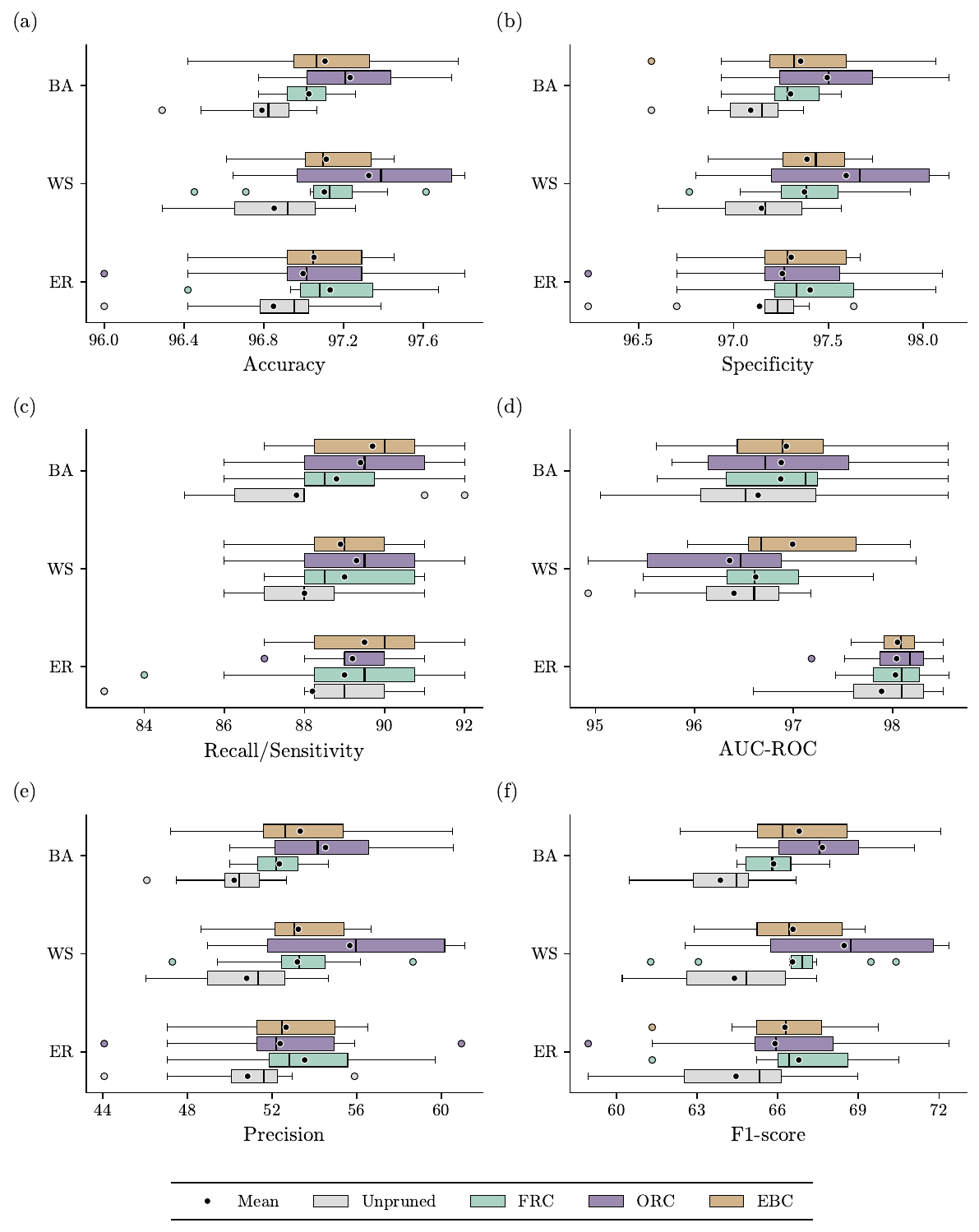}
\caption{Comparison of the three classes of RWNNs, namely ER, WS, and BA, under four scenarios: before pruning and after pruning based on three edge-centric network measures, FRC, ORC, and EBC, across the six performance metrics: (a) accuracy, (b) specificity, (c) recall or sensitivity, (d) AUC-ROC, (e) precision, and (f) F1-score.}
\label{fig_Measures_comparison}
\end{figure*}

For each RWNN class, performances are reported as percentages for all three edge-centric measures. 
Table \ref{tab_pruned} presents the average and maximum values, while figures \ref{fig_Measures_comparison}(a)-(f) display the corresponding box plots across the six performance metrics.

In the ER class of RWNNs, overall, all three measures, FRC, ORC, and EBC, achieved high accuracy, with mean values ranging from 96.997 for ORC to 97.132 for FRC (see figure \ref{fig_Measures_comparison}(a) and table \ref{tab_pruned}). 
ORC achieved the highest maximum accuracy of 97.806, closely followed by FRC at 97.677, while EBC records 97.452. 
In terms of specificity, FRC recorded the best average value of 97.403, whereas EBC and ORC follow closely with 97.303 and 97.257, respectively. 
However, ORC achieved the best maximum value of 98.1, while FRC and EBC achieved 98.067 and 97.667, respectively.
Sensitivity scores are comparable across all three measures, with mean values around 89 to 89.5 and maximum values ranging from 91 to 92. 
EBC attained the highest mean sensitivity of 89.5, while both FRC and EBC reached the best maximum value of 92.
In terms of AUC-ROC, the scores are comparable across all three measures. 
EBC recorded the highest mean value of 98.051, while FRC achieved the best maximum value of 98.567. 
For Precision, FRC attained the highest average of 53.543, whereas ORC reached the best maximum of 60.959. 
A similar pattern is observed for the F1-score, where FRC provided the best average of 66.781 and ORC delivered the highest maximum of 72.358.
The findings highlight that all three edge-centric measures achieve high performance for most of the performance metrics on the ER class of RWNNs.
FRC yields the most consistent average performance across multiple metrics, including accuracy, specificity, precision, and F1-score. 
In contrast, in most cases, ORC attains the highest maximum values, highlighting its potential to capture peak performance.
EBC, meanwhile, stands out in terms of mean sensitivity and AUC-ROC.
Collectively, these results indicate complementary strengths across the measures, with FRC offering stability, ORC excelling in extreme cases, and EBC providing advantages for sensitivity and AUC-ROC.

In the WS class of RWNNs, all three measures achieve very high and comparable accuracy. 
However, ORC achieved the best mean accuracy of 97.326 and also the highest maximum accuracy of 97.806. 
Thus, ORC demonstrated a clear advantage in accuracy, although the differences are relatively small among the edge-centric measures (see table \ref{tab_pruned}).
Specificity values are also consistently high across measures. 
ORC again achieved the highest mean value of 97.593 and the maximum value of 98.133.
Here, ORC shows superiority, while FRC and EBC perform very closely to one another.
In terms of sensitivity, the three measures yield comparable results, with only slight variations (see table \ref{tab_pruned}). 
ORC performed best, with a mean of 89.3 and a maximum of 92. 
For AUC-ROC, performance is again strong and closely aligned across measures. 
EBC achieved the highest mean value at 96.992, while ORC reached the best maximum at 98.238 but has the lowest mean among the three at 96.355. 
In terms of precision, ORC achieved the highest mean of 55.677 and the maximum of 61.111.
A similar trend is seen in the F1-score, where ORC again achieved the best performance with a mean of 68.47 and a maximum of 72.358.
Overall, the WS class demonstrates strong and consistent performance across all three edge-centric measures. 
ORC stands out, achieving the highest values in accuracy, specificity, precision, and F1-score, while also maintaining the best sensitivity among the three measures. 
FRC performs reliably and reaches competitive maximum values, though it generally falls just short of ORC. 
EBC, on the other hand, shows the best average AUC-ROC, even though its precision and F1-scores are lower. 
Collectively, these results suggest that ORC is the most effective measure for the WS class.

The BA class also shows similar performance across all three edge-centric measures (see figure \ref{fig_Measures_comparison} and table \ref{tab_pruned}).
In terms of accuracy, ORC achieved the highest mean of 97.232, while EBC records the highest maximum at 97.774. 
In specificity, ORC again led with the best mean of 97.493 and maximum of 98.133, followed by EBC and then FRC. 
The sensitivity scores are comparable overall.
EBC achieved the best mean at 89.7, and each of the three measures reached the maximum of 92.
AUC-ROC values are very close across the measures, with EBC attaining the highest mean at 96.927 and all three measures achieving the same maximum at 98.561. 
In terms of precision, ORC was strongest with the best mean at 54.527 and maximum at 60.563. 
Finally, ORC delivered the highest mean F1-score at 67.658, while EBC attained the best maximum at 72.065.
Overall, in the BA class, ORC stands out with the best average performance, especially in accuracy, specificity, precision, and F1-score. 
EBC, on the other hand, excels in recall and attains the strongest maximum F1-score. 
Although FRC remains steady and yields results similar to ORC and EBC, it generally falls slightly behind in overall performance.
Together, these findings highlight ORC as the most effective edge-centric measure for the BA class.

%---------------------------------------------------------------
% Table 7
\begin{table}
\caption{Performance summary in terms of accuracy, specificity, sensitivity, AUC-ROC, precision, and F1-score, for pruned RWNN configurations across three classes.}
\label{tab_pruned}
\begin{tabular*}{\columnwidth}{@{\extracolsep\fill}llccccccccccccc@{\extracolsep\fill}}
\toprule 
\toprule
\multirow{2}{*}{\textbf{Model}} & \multirow{2}{*}{\shortstack[l]{\textbf{Edge}\\\textbf{measure}}} & \multicolumn{2}{c}{\textbf{Accuracy (\%)}} & \multicolumn{2}{c}{\textbf{Specificity (\%)}} & \multicolumn{2}{c}{\textbf{Sensitivity (\%)}} & \multicolumn{2}{c}{\textbf{AUC-ROC (\%)}} & \multicolumn{2}{c}{\textbf{Precision (\%)}} & \multicolumn{2}{c}{\textbf{F1-score (\%)}} \\
\cline{3-4} \cline{5-6} \cline{7-8} \cline{9-10} \cline{11-12} \cline{13-14}
& & Average & Max & Average & Max & Average & Max & Average & Max & Average & Max & Average & Max\\
\hline
\multirow{3}{*}{\textbf{ER}} & FRC & 97.132 & 97.677 & 97.403 & 98.067 & 89 & 92 & 98.03 & 98.567 & 53.543 & 59.722 & 66.781 & 70.492 \\
& ORC & 96.997 & 97.806 & 97.257 & 98.1 & 89.2 & 91 & 98.041 & 98.51 & 52.381 & 60.959 & 65.896 & 72.358 \\
& EBC & 97.052 & 97.452 & 97.303 & 97.667 & 89.5 & 92 & 98.051 & 98.51 & 52.663 & 56.522 & 66.265 & 69.732 \\
\midrule
\multirow{3}{*}{\textbf{WS}} & FRC & 97.103 & 97.613 & 97.373 & 97.933 & 89 & 91 & 96.619 & 97.804 & 53.195 & 58.667 & 66.545 & 70.4 \\
& ORC & 97.326 & 97.806 & 97.593 & 98.133 & 89.3 & 92 & 96.355 & 98.238 & 55.677 & 61.111 & 68.47 & 72.358 \\
& EBC & 97.113 & 97.452 & 97.387 & 97.733 & 88.9 & 91 & 96.992 & 98.177 & 53.243 & 56.688 & 66.565 & 69.261 \\
\midrule
\multirow{3}{*}{\textbf{BA}} & FRC & 97.026 & 97.258 & 97.3 & 97.567 & 88.8 & 92 & 96.872 & 98.561 & 52.344 & 54.658 & 65.842 & 67.925 \\
& ORC & 97.232 & 97.742 & 97.493 & 98.133 & 89.4 & 92 & 96.877 & 98.561 & 54.527 & 60.563 & 67.658 & 71.074 \\
& EBC & 97.106 & 97.774 & 97.353 & 98.067 & 89.7 & 92 & 96.927 & 98.561 & 53.332 & 60.544 & 66.793 & 72.065 \\
\bottomrule
\end{tabular*}
\end{table}
%---------------------------------------------------------------

%---------------------------------------------------------------
% Figure 2
%---------------------------------------------------------------

Across the three RWNN classes (ER, WS, and BA), all edge-centric measures provided consistently strong performance with distinct strengths. 
ORC demonstrated highly competitive performance overall, achieving the best accuracy, specificity, precision, and F1-score in the WS and BA classes, as well as some of the highest maximum values in ER class. 
EBC, although generally weaker in precision and F1-score, achieved strong performance in sensitivity and AUC-ROC. 
FRC demonstrated stable and competitive results, achieving the best average performance in the ER class and remaining comparable in WS and BA classes.
Taken together, these findings highlight ORC as the most effective and consistent measure across all RWNN classes, FRC offers stable averages, and EBC offers complementary strengths in sensitivity and AUC-related performance.
A summary of the confusion matrix components across all evaluated instances is provided in table \ref{tab_confusion}, which also indicates that the different edge-centric measures achieve largely comparable results for different instances.

%---------------------------------------------------------------

%---------------------------------------------------------------
\subsection{Pruning potential of the edge-centric network measures}

In this subsection, we report the pruning potential of each of the three edge-centric network measures on each of the three classes of RWNNs in terms of compression ratio, percentage of parameters pruned, and theoretical speedup achieved.

%---------------------------------------------------------------
\subsubsection{Compression ratios of the pruned networks} \label{subsubsec_ComRatioPruned}

The compression ratio is defined as the ratio of the size of the original network to that of the pruned network, where network size is measured in terms of the number of parameters. 
Importantly, we did not predefine compression ratios prior to pruning. 
Instead, using a binary search, we aimed to obtain the smallest possible versions of the initial RWNNs that still match baseline performance, and then evaluated their compression ratio. 
We further emphasize that the optimal compressed version is regarded as the smallest network that preserves the original performance, rather than the version that achieves the highest accuracy, specificity, or sensitivity.

\begin{figure*}
\centering
\includegraphics[width=12cm]{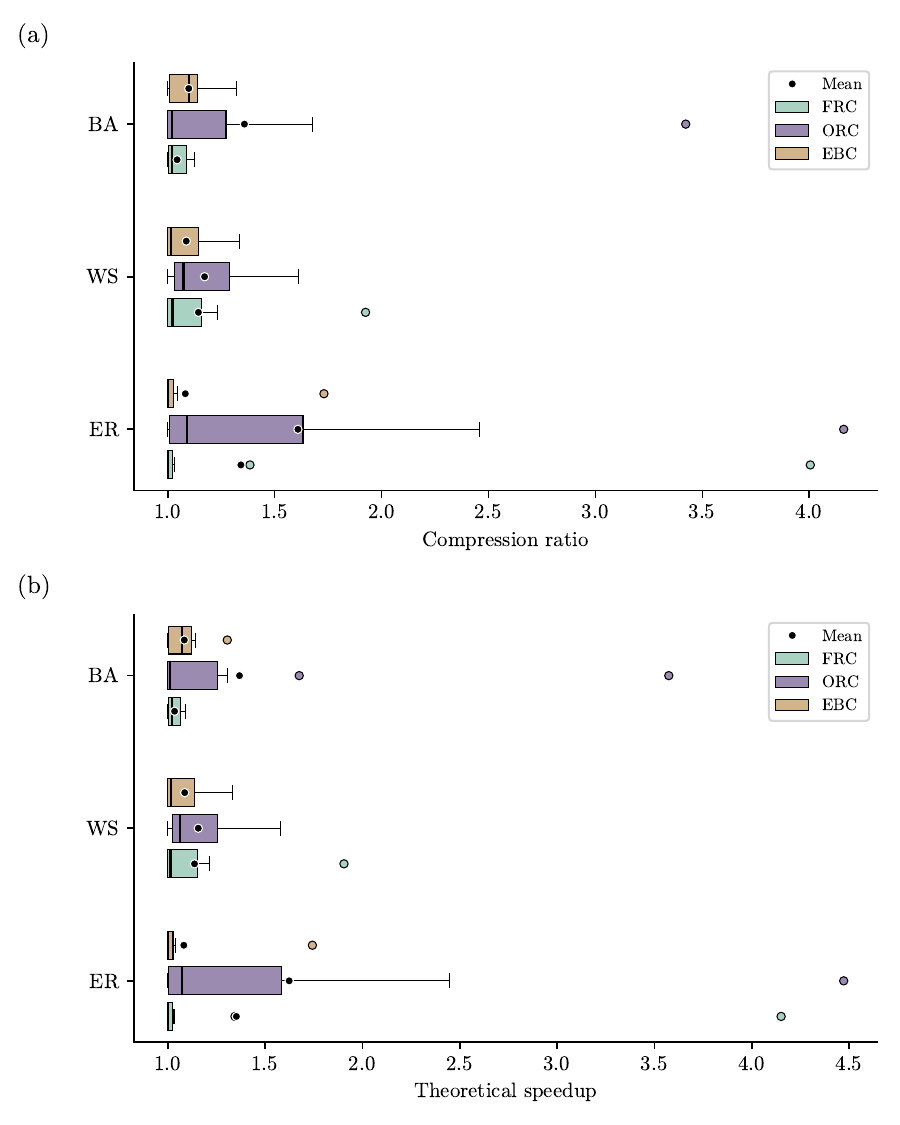}
\caption{Pruning performance in terms of compression ratio and theoretical speedup for each of the three edge-centric network measures for pruned RWNN configurations across three classes namely ER, WS, and BA.}
\label{fig_PruningPotential}
\end{figure*}

In the ER class of RWNNs, ORC yielded the highest compression ratio with a maximum of 4.161 and an average of 1.609. 
FR achieved the second-highest compression ratio with a maximum of 4.005 and an average of 1.342, while EBC achieved a maximum of 1.731 and an average of 1.082.
Overall, ORC achieved a higher compression ratio than the other two measures (see figure \ref{fig_PruningPotential}(a)). 
In terms of the percentage of parameters retained, ORC achieved a minimum of 24.033\%, FRC of 24.97\%, and EBC of 57.785\% (see table \ref{tab_compreRatio}). 
These results indicate that both curvature-based measures, FRC and ORC, exhibit comparable pruning potential for the ER class of RWNNs.

In the WS class of RWNNs, FRC achieved the highest compression ratio with a maximum of 1.925 and an average of 1.144. 
ORC followed with a maximum of 1.611 and a slightly higher average of 1.172, while EBC achieved a compression ratio of a maximum of 1.337 and an average of 1.087. 
Thus, FRC achieves the highest compression ratio, while ORC yields the highest average compression ratio (see figure \ref{fig_PruningPotential}(a) and table \ref{tab_compreRatio}). 
In terms of the percentage of parameters retained, FRC reached a minimum of 51.945\%, ORC of 62.058\%, and EBC of 74.81\% (see table \ref{tab_compreRatio}). 
Overall, in the WS class of RWNNs, FRC achieves the maximum pruning, while ORC is more effective on average.

In the BA class of RWNNs, ORC achieved the highest compression ratio with a maximum of 3.422 and an average of 1.359; EBC achieved the second highest compression ratio with 1.321 and an average of 1.098.
FRC achieved a maximum compression of 1.124 and an average of 1.044 (see figure \ref{fig_PruningPotential}(a)). 
In terms of the percentage of the parameters retained, ORC had a minimum of 29.219\%, compared to 75.685\% for EBC and 88.976\% for FRC (see table \ref{tab_compreRatio}). 
These results indicate that ORC has better pruning potential for the BA class of RWNNs relative to the other two edge-centric measures.

In a global comparison, our results show that curvature-based measures (FRC and ORC) generally provide better pruning performance than edge betweenness centrality, a standard network metric.
When comparing between the two curvature-based measures, ORC consistently achieves the highest compression ratios across the ER and BA classes, while FRC provides competitive performance in the ER and WS classes. 

%---------------------------------------------------------------
% Figure 3
%---------------------------------------------------------------
%---------------------------------------------------------------
% Table 8
\begin{table}
\caption{Pruning performance in terms of compression ratio, percentage of parameters retained, and theoretical speedup for each of the three edge-centric network measures for pruned RWNN configurations across three classes namely ER, WS, and BA.}
\label{tab_compreRatio}
\begin{tabular*}{0.9\columnwidth}
{@{\extracolsep\fill}llccccccc@{\extracolsep\fill}}
\toprule 
\toprule 
\multirow{2}{*}{\textbf{Model}} & \multirow{2}{*}{\shortstack[l]{\textbf{Edge}\\ \textbf{measure}}} & \multicolumn{2}{c}{\textbf{Compration ratio}} & \multicolumn{2}{c}{\textbf{Parameters retained (\%)}} & \multicolumn{2}{c}{\textbf{Theoretical speedup }} \\
\cline{3-4} \cline{5-6} \cline{7-8}
& & Average & Max & Average & Min & Average & Max \\
\hline
\multirow{3}{*}{\textbf{ER}} & FRC & 1.342 & 4.005 & 89.404 & 24.97 & 1.353 & 4.151 \\
& ORC & 1.609 & 4.161 & 76.824 & 24.033 & 1.624 & 4.472 \\
& EBC & 1.082 & 1.731 & 94.876 & 57.785 & 1.083 & 1.743 \\
\midrule
\multirow{3}{*}{\textbf{WS}} & FRC & 1.144 & 1.925 & 90.882 & 51.945 & 1.138 & 1.905 \\
& ORC & 1.172 & 1.611 & 87.353 & 62.058 & 1.157 & 1.577 \\
& EBC & 1.087 & 1.337 & 92.955 & 74.81 & 1.087 & 1.334 \\
\midrule
\multirow{3}{*}{\textbf{BA}} & FRC & 1.044 & 1.124 & 95.961 & 88.976 & 1.035 & 1.091 \\
& ORC & 1.359 & 3.422 & 84.932 & 29.219 & 1.369 & 3.574 \\
& EBC & 1.098 & 1.321 & 91.694 & 75.685 & 1.085 & 1.306 \\
\bottomrule
\end{tabular*}
\end{table}
%---------------------------------------------------------------

%---------------------------------------------------------------
\subsubsection{Theoretical speedup of the pruned networks}

Theoretical speedup is defined as the ratio of the total number of FLOPs required by the original network to the FLOPs required by the pruned network. 
This measure provides an approximate indication of the computational advantage gained through pruning, reflecting the potential reduction in overall computation cost during both training and inference.

We observed that the theoretical speedup closely follows the same trend as the compression ratio in determining which measure performs best for each class of RWNNs. 
In the ER class, ORC achieved the highest speedup, reaching a maximum of 4.472 with an average of 1.624. 
For the WS class, FRC attained the maximum speedup of 1.905, with an average of 1.138, while ORC provided a slightly higher average speedup of 1.157 and a maximum of 1.577. 
In the BA class, ORC again showed the better performance, with a maximum speedup of 3.574 and an average of 1.369. 
A detailed comparison of these results is presented in table \ref{tab_compreRatio}, and a visual representation is provided in figure \ref{fig_PruningPotential}(b).
%---------------------------------------------------------------

%---------------------------------------------------------------
% Figure 4
\begin{figure*}
\centering
\includegraphics[width=\linewidth]{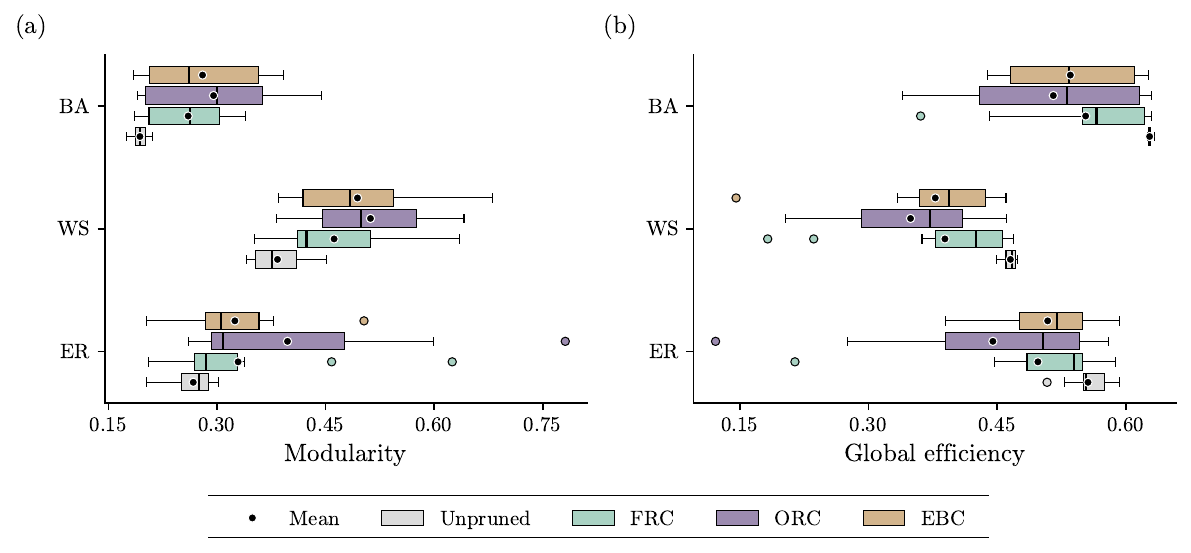}
\caption{Comparison of the three classes of RWNNs, namely ER, WS, and BA, under four scenarios: before pruning and after pruning with FRC, ORC, and EBC, evaluated using two network-based measures: (a) modularity and (b) global efficiency.}
\label{fig_modularity}
\end{figure*}
%---------------------------------------------------------------

%---------------------------------------------------------------

\subsection{Pruning leads to increase in modularity and decrease in global efficiency of the underlying random network}

In this section, we investigate how the structure of the neural networks changed after pruning based on each of the three edge-centric measures. 
For this purpose, we considered two global network measures: (a) modularity \cite{newman2006} and (b) global efficiency \cite{latora2001} of the network.
For this analysis, the network measures were computed using the \textit{NetworkX} library \cite{hagberg2008} in Python.

\textit{Modularity} evaluates the strength of division of a network into communities (or modules). 
It quantifies how well a network is partitioned compared to a random baseline.
We considered the greedy modularity maximization algorithm \cite{clauset2004} to partition the network into communities. 
The resulting partitions were then utilized to compute the modularity value $Q$, which is defined as:
\begin{equation*}
        Q = \frac{1}{2m} \sum_{i,j} \left[ A_{ij} - \frac{k_i k_j}{2m} \right] \delta(c_i, c_j)
\end{equation*}
where $m$ is the number of edges, $A$ is the adjacency matrix, $k_i$ is the degree of the node $i$, and $\delta(c_i, c_j)$ is 1 if both the nodes $i$ and $j$ are in the same community, else 0.

\textit{Global efficiency} quantifies how efficiently information is transferred over a network. 
It is calculated as the average of the inverse shortest path lengths between all pairs of nodes.
For a graph G with N nodes, global efficiency is defined as,
\begin{equation*}
        E(G) = \frac{1}{N(N-1)} \sum_{i \neq j} \frac{1}{d_{ij}}
\end{equation*}
where $d_{ij}$ is the shortest path length between nodes $i$ and $j$.
If nodes $i$ and $j$ belong to different connected components, i.e., no path exists between them, then $\tfrac{1}{d_{ij}}$ becomes 0.

Before pruning, the modularity values among the three RWNN classes were highest for WS, followed by ER and BA. 
After pruning, modularity increased consistently across all three pruning methods (FRC, ORC, and EBC) across all classes of RWNNs (see figure \ref{fig_modularity}(a) and table \ref{tab_modularity}). 
Among them, pruning based on ORC achieved the highest average modularity across all three classes of RWNNs. 
Furthermore, ORC yielded the maximum modularity in the ER and BA classes, with values of 0.781 and 0.444, respectively. 
In contrast, for the WS class, the highest modularity of 0.681 was achieved with EBC-based pruning.

In terms of global efficiency, we observed the reverse trend (see figure \ref{fig_modularity}(b) and table \ref{tab_modularity}). 
This aligns with Baum \textit{et al.} \cite{baum2017}, who report a negative correlation between modularity and global efficiency, suggesting that stronger modular segregation often reduces efficiency.
Nonetheless, Song \textit{et al.} \cite{song2014} observed relatively stable efficiency values across varying levels of modularity, whereas Romano \textit{et al.} \cite{romano2018} demonstrated a non-linear association in which global efficiency peaked at intermediate modularity levels before declining at higher values.
In this study, for the unpruned RWNNs, global efficiency was lowest for WS, followed by ER, and highest for BA.
Notably, after pruning, global efficiency decreased consistently for all three pruning approaches (FRC, ORC, and EBC) across all classes of RWNNs. 
Among them, pruning based on ORC achieved the lowest average global efficiency across all three classes of RWNNs.
The highest average global efficiency was obtained with FRC in the WS and BA classes, while EBC yielded the best average in the ER class. 
In terms of maximum global efficiency, all the pruning measures achieved nearly similar values across the three RWNNs.
Table \ref{tab_modularity} summarizes the average and maximum values of modularity and global efficiency across each class of RWNNs before and after pruning.

Overall, pruning produced opposite effects on modularity and global efficiency across RWNNs. 
Modularity consistently increased after pruning, with ORC generally yielding the highest average values. 
By contrast, global efficiency consistently declined across all pruning methods, with ORC showing the lowest averages. 
These findings highlight a trade-off, where pruning strengthens modular segregation but reduces efficiency.
This pattern reflects the inverse relationship between modularity and global efficiency often reported in complex network studies \cite{kim2014, tosh2015, baum2017}.

%---------------------------------------------------------------

%---------------------------------------------------------------
% Table 9
\begin{table}
\caption{Summary of modularity and global efficiency for the three classes of RWNNs before pruning (unpruned) and after pruning based on three edge-centric network measures.}
\label{tab_modularity}
\begin{tabular*}{0.7\columnwidth}
{@{\extracolsep\fill}llccccc@{\extracolsep\fill}}
\toprule 
\toprule 
\multirow{2}{*}{\textbf{Model}} & \multirow{2}{*}{\shortstack[l]{\textbf{Edge}\\ \textbf{measure}}} & \multicolumn{2}{c}{\textbf{Modularity}} & \multicolumn{2}{c}{\textbf{Global efficiency}} \\
\cline{3-4} \cline{5-6}
& & Average & Max & Average & Max \\
\midrule
\multirow{4}{*}{\textbf{ER}} & Unpruned & 0.267 & 0.302 & 0.556 & 0.592 \\
& FRC & 0.329 & 0.625 & 0.497 & 0.588 \\
& ORC & 0.397 & 0.781 & 0.445 & 0.58 \\
& EBC & 0.325 & 0.503 & 0.509 & 0.592 \\
\midrule
\multirow{4}{*}{\textbf{WS}} & Unpruned & 0.384 & 0.452 & 0.465 & 0.474 \\
& FRC & 0.462 & 0.634 & 0.389 & 0.469 \\
& ORC & 0.512 & 0.641 & 0.349 & 0.46 \\
& EBC & 0.494 & 0.681 & 0.378 & 0.46 \\
\midrule
\multirow{4}{*}{\textbf{BA}} & Unpruned & 0.193 & 0.211 & 0.628 & 0.633 \\
& FRC & 0.26 & 0.34 & 0.553 & 0.63 \\
& ORC & 0.295 & 0.444 & 0.515 & 0.63 \\
& EBC & 0.28 & 0.392 & 0.535 & 0.626 \\
\bottomrule
\end{tabular*}
\end{table}
%---------------------------------------------------------------

%---------------------------------------------------------------
\section{Conclusion and limitations}

This work evaluated the use of discrete Ricci curvature-based measures for pruning randomly wired neural networks (RWNNs) in the classification of COVID-19 images. 
Glass \textit{et al.} \cite{Glass2020} introduced \textit{RicciNets}, which leverage Ollivier-Ricci curvature (ORC) to prune randomly wired neural networks by preserving the most salient computational paths. 
Building on this idea, our work incorporates Forman-Ricci curvature (FRC) and edge betweenness centrality (EBC) alongside ORC. 
We focus particularly on evaluating whether FRC, which is computationally more efficient, can serve as a practical alternative while maintaining comparable pruning performance.
In \textit{RicciNets}, the authors have used the Watts-Strogatz (WS) random graph model with $k=4$ as the network generator for RWNNs, whereas this study considered three different classical random graph models, namely, Erd\"{o}s-R\'{e}nyi (ER), Watts-Strogatz (WS), and Barab\'{a}si-Albert (BA). 
For each of the three RWNN classes, we generated ten distinct network instances by considering ten separate random seeds.
By evaluating ER, WS, and BA graph structures, we demonstrated that pruning guided by FRC, ORC, and EBC can reduce network complexity while maintaining strong performance.
Among these, FRC offered consistent average results, while ORC often achieved peak performance. 
These findings suggest that curvature-driven pruning provides a principled alternative to random or magnitude-based pruning approaches.
On the other hand, EBC provides complementary advantages by enhancing sensitivity and AUC-related outcomes.
Furthermore, pruning enhances the modularity of RWNNs, resulting in a more structured organization; however, this improvement is accompanied by a decline in the global efficiency of the underlying random network.
Our findings suggest that discrete Ricci curvatures offer a principled, geometry-inspired approach to network pruning, complementing existing methods such as weight and filter pruning in CNNs. 
While CNN pruning is often based on heuristics such as weight magnitude, curvature-based measures capture structural importance at the graph level, potentially enabling more efficient pruning without compromising generalization.
Although recent advances in large language models (LLMs) and generative AI dominate current research, they also face pressing challenges of efficiency and scalability. 
The insights gained here demonstrate that discrete Ricci curvatures can prune redundant connections while maintaining model performance, highlighting a direction that may translate to transformer-based architectures, offering avenues for reducing energy consumption and improving interpretability in future foundation models.

While our study demonstrates the potential of Ricci curvature-based pruning in randomly wired neural networks (RWNNs), several limitations should be acknowledged. 
The dataset was notably imbalanced, with far fewer positive cases compared to negatives. 
To address this, augmentation was employed to enrich the minority class and reduce the likelihood of overfitting. 
Although the consistent performance across repeated runs suggests that this strategy was effective, the possibility of residual bias cannot be entirely ruled out. 
In addition, the experiments were conducted under limited computational resources, primarily using Google Colab Pro. 
This placed constraints on the scale of training, hyperparameter tuning, and the number of repeated trials that could be performed. 
Our analysis also focused exclusively on RWNNs, without extending curvature-driven pruning to conventional CNNs, where pruning learned weights could provide a valuable point of comparison. 

In summary, our study provides initial evidence that curvature-guided pruning can effectively simplify neural networks while retaining competitive performance. 
Future work should extend this framework to larger datasets, diverse modalities, and more standard deep architectures, enabling a rigorous comparison with conventional pruning strategies.

%---------------------------------------------------------------

%---------------------------------------------------------------
\subsection*{COMPETING INTERESTS}
The authors declare no competing interests.

\subsection*{AUTHOR CONTRIBUTIONS}
Designed the research: P.E., S.V., M.M., A.S.; Performed the research: P.E., S.V., M.M., A.S.; Performed the computations: P.E., S.V., M.M.; Wrote the paper: P.E., S.V., M.M., A.S.

\subsection*{ACKNOWLEDGMENTS}
We thank Sarath Jyotsna Ramaia for technical support and discussion during the initial phase of this project.
A.S. acknowledges funding from the Department of Atomic Energy, Government of India (via the Apex project to The Institute of Mathematical Sciences (IMSc), Chennai) and funding from the Max Planck Society, Germany (through the award of a Max Planck Partner Group).

\subsection*{DATA AND CODE AVAILABILITY}
All the necessary data and codes are deposited in GitHub to reproduce the results in this manuscript, and are available at: \url{https://github.com/asamallab/CurvBased_RWNNPrune}.
%---------------------------------------------------------------

%---------------------------------------------------------------
% References
\bibliographystyle{unsrt}
\bibliography{arXiv/refs}
%---------------------------------------------------------------

%---------------------------------------------------------------
\end{document}